\documentclass[10pt,conference]{IEEEtran}
\IEEEoverridecommandlockouts

\usepackage{cite}
\usepackage{booktabs}

\usepackage{amsmath,amssymb,amsfonts,amsthm}

\usepackage[ruled,linesnumbered]{algorithm2e}
\SetKwComment{Comment}{/* }{ */}

\usepackage{fancyhdr}
\theoremstyle{definition}

\usepackage{caption}
\usepackage{graphicx}
\usepackage{textcomp}
\usepackage{xcolor}
\usepackage{tikz}

\def\BibTeX{{\rm B\kern-.05em{\sc i\kern-.025em b}\kern-.08em
T\kern-.1667em\lower.7ex\hbox{E}\kern-.125emX}}

\usepackage{hyperref}
\hypersetup{
    colorlinks=true,
    linkcolor=blue,
    filecolor=magenta,      
    urlcolor=cyan,
    }
\usepackage{multicol}
\usepackage{multirow}

\usepackage{listings}

\lstdefinestyle{mystyle}{
    backgroundcolor=\color{backcolour},   
    commentstyle=\color{codegreen},
    keywordstyle=\color{magenta},
    numberstyle=\tiny\color{codegray},
    stringstyle=\color{codepurple},
    basicstyle=\ttfamily\footnotesize,
    breakatwhitespace=false,         
    breaklines=true,                 
    captionpos=b,                    
    keepspaces=true,                 
    numbers=left,                    
    numbersep=5pt,                  
    showspaces=false,                
    showstringspaces=false,
    showtabs=false,                  
    tabsize=2
}
\usepackage{enumitem}
\setlist[itemize]{align=parleft,left=0pt..1em}
\usepackage[normalem]{ulem}
\usepackage{framed}
\definecolor{shadecolor}{gray}{0.9}

\newcommand*{\affaddr}[1]{#1} 

\newcommand*{\email}[1]{\texttt{#1}}

\begin{document}

\title{Toward a Predictive eXtended Reality Teleoperation System with Duo-Virtual Spaces}

\author{%
Ziliang Zhang, Cong Liu, Hyoseung Kim\\
\affaddr{University of California, Riverside}\\
\email{\{zzhan357, congl, hyoseung\}@ucr.edu}%
}
 
\maketitle

\thispagestyle{fancy}

\begin{abstract} 
Extended Reality (XR) provides a more intuitive interaction method for teleoperating robots compared to traditional 2D controls. Recent studies have laid the groundwork for usable teleoperation with XR, but it fails in tasks requiring rapid motion and precise manipulations due to the large delay between user motion and agent feedback. 
In this work, we profile the end-to-end latency in a state-of-the-art XR teleoperation system and propose our idea to optimize the latency by implementing a duo-virtual spaces design and localizing the agent and objects in the user-side virtual space, while calibrating with periodic ground-truth poses from the agent-side virtual space.
\end{abstract}

\section{Introduction}

Extended Reality (XR) has proven its strength in intuitive robot control as it displays the robot movement directly within an XR-maintained virtual space that can be easily comprehended by common users~\cite{whitney2019comparing,hetrick2020comparing,karpichev2024extended}. Emergent research enables the teleoperation of robots with XR, which performs remote human-robot interaction through WebSocket client-server design similar to existing teleconferencing applications. This enables risky operations in a hazardous environment or inaccessible locations like fire scenes or outer space~\cite{whitney2018ros,rossharp}.

\begin{figure}[t]
\includegraphics[width=\linewidth]{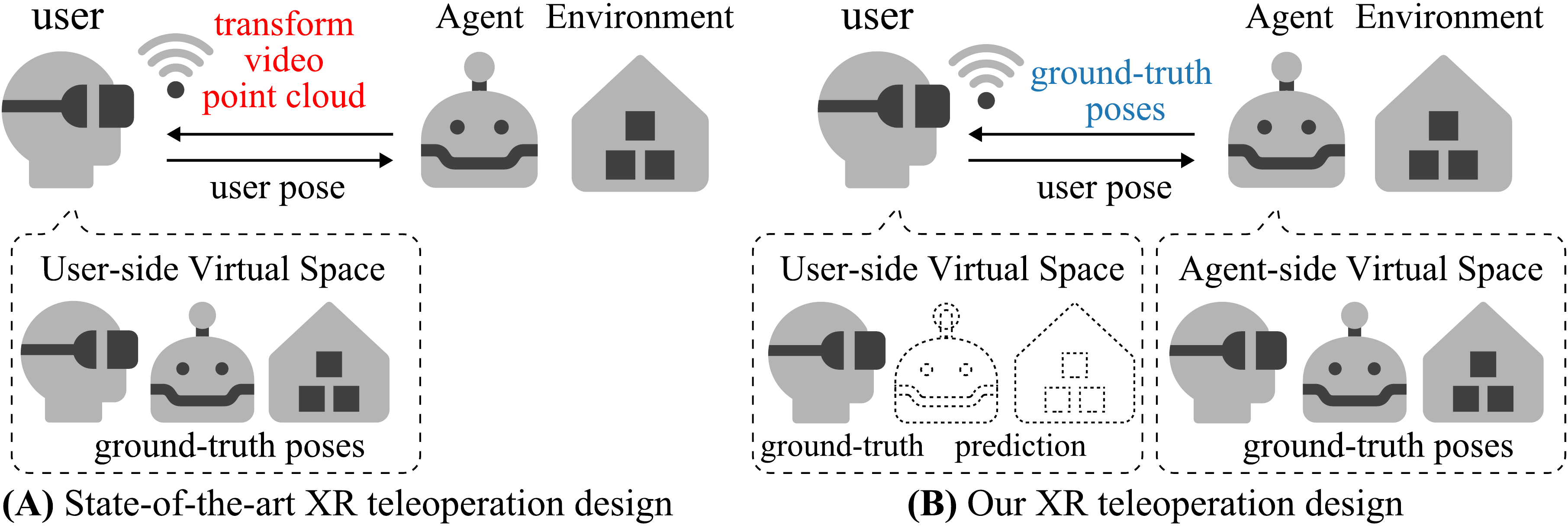}
\caption{Comparison between existing teleoperation and our proposed system. Our design conducts local prediction in XR for remote agent and objects, mitigating large network delay. }
\label{fig_intuition}
\end{figure}

Existing XR teleoperation constructs a single virtual space on the user side for interaction, as shown in Fig.~\ref{fig_intuition}(A). This user-side virtual space consists of user poses provided by XR sensors, as well as agent pose and objects poses provided by the robot agent in the remote working environment. 
Since these poses are reliably captured by on-device sensors without additional processing after localization, we denote them as ground-truth poses. 
The user head, body, and hand poses captured by XR sensors are initially transmitted via the internet to the agent to perform coordinate transformations. During the transform, the agent sends transform string, video feed, and point cloud information back to the XR for virtual space visualization, which is displayed to the user through a head-mount display.
However, since the virtual space must rely on remote agent and objects poses to update for every output frame, significant network latency causes disjoint user and agent movements, which leads to tasks with rapid maneuvers or precise manipulations unable to be completed~\cite{whitney2018ros}. 

To solve this problem, we propose a duo-virtual spaces design that localizes the agent and objects through the use of XR sensors and object modeling to avoid the large latency caused by communication. Our design is illustrated in Fig.~\ref{fig_intuition}(B). The preliminary results from our case study show that by using this design, the end-to-end latency of XR-robot teleoperation can be reduced by up to 89\%.

\section{Background and Motivation}

\begin{figure*}[t]
\includegraphics[width=\textwidth]{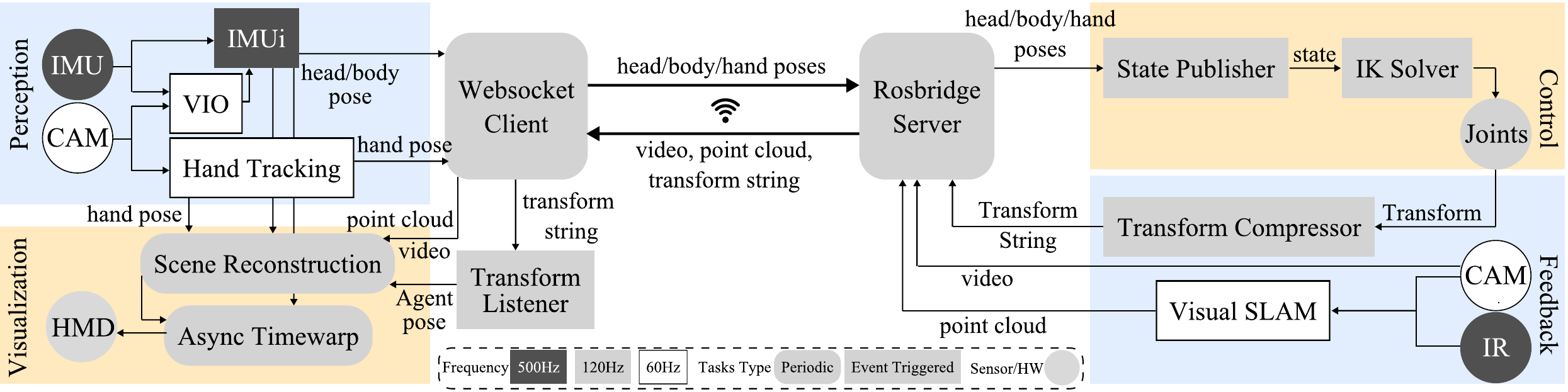}
\caption{ROS Reality~\cite{whitney2018ros}, an open-sourced XR teleoperation system, implements a publisher-subscriber execution model for XR and uses ROS for agent. Agent depends on user poses to transform and XR depends on video, point cloud, and transform string to update the virtual space.}
\label{fig_system_overview}
\vspace{-1mm}
\end{figure*}

\subsection{Characterizing Teleoperation Workload}

State-of-the-art XR teleoperation maintains a single virtual space in the XR framework that consists of user, agent, and objects poses to localize agent and objects and display them to the user. The detailed workflow is explained below with ROS Reality~\cite{whitney2018ros}, a representative XR teleoperation framework widely used for prevalent commercial products and projects like ROS Sharp~\cite{rossharp} and Open-TeleVision~\cite{cheng2024open}. Fig.~\ref{fig_system_overview} shows its detailed system design and task dependencies.

The XR framework follows a publisher-subscriber model for task execution. The current XR devices have multiple CPUs and accelerators (GPU, NPUs) for running tasks concurrently.
The XR system reads from an inertial measurement unit (IMU) and a camera (CAM) sensor and estimates user head and body poses through Visual Inertial Odometer (VIO) and IMU integration (IMUi) tasks. Concurrently, a hand-tracking model takes the CAM input and produces the hand poses at the same rate as the body pose. All poses are sent through a WebSocket client to the agent side, in return for the video, point cloud, and transform string. Before visualizing the virtual space, a transform listener task is used to convert the transform string to agent pose, and point cloud is used to update the objects poses. During visualization, the scene reconstruction task constructs the state for virtual space based on the most recent head, body, and hand poses from the XR perception phase, and localizes the agent and objects using the received poses. Finally, an Asynchronous Timewarp task is scheduled immediately before frame submission so XR can recalculate the viewport based on the head pose that has the smallest possible age of information.

The agent uses the Robot Operating System (ROS) for scheduling and executing tasks on agent. It first maintains a Rosbridge server to receive the user poses that include head, body, and hand poses. Upon receiving these user poses, the agent implements a state publisher to convert the poses to state for the robot Inverse Kinematic (IK) solver. The IK solver outputs the transform command to the robot's joints and starts transforming. During the process, a transform compress task constantly compresses the transform command to a transform string. Concurrently, a visual SLAM task takes input from agent CAM and infrared (IR) sensors and produces point cloud data. The transform string, point cloud, and the video feed directly from CAM are constantly sent back to the user side via the Rosbridge server. 


\vspace{-1mm}
\subsection{Challenges of latency optimization}

Although ROS reality enables basic usability, tasks like tossing and catching balls, tying shoe laces, and tracing straight lines are still unachievable due to their rapid objects movement or precise manipulation requirements~\cite{whitney2018ros}. This is due to the large latency between a user motion and the visual feedback from the agent in the virtual space~\cite{hetrick2020comparing,karpichev2024extended}. To profile the latency, we set up an experiment with two PCs running the XR framework and a Gazebo simulation of Kinova J2N6S300~\cite{kinova} robot arm in two different network segmentations from the same building. The PCs communicate through a 5GHZ 802.11ac WIFI environment. During the experiment, the XR framework follows three trajectories in EuRoC MAV dataset~\cite{euroc_mav} with two different scenes and sent poses at the same frequency as the screen refresh rate of the head-mount display. We use the ROS reality~\cite{whitney2018ros} framework for all data communication and record the age of information of the user pose that leads to a transform when frame is submitted to head-mount display as end-to-end latency.

\begin{figure}[t]
\includegraphics[width=\linewidth]{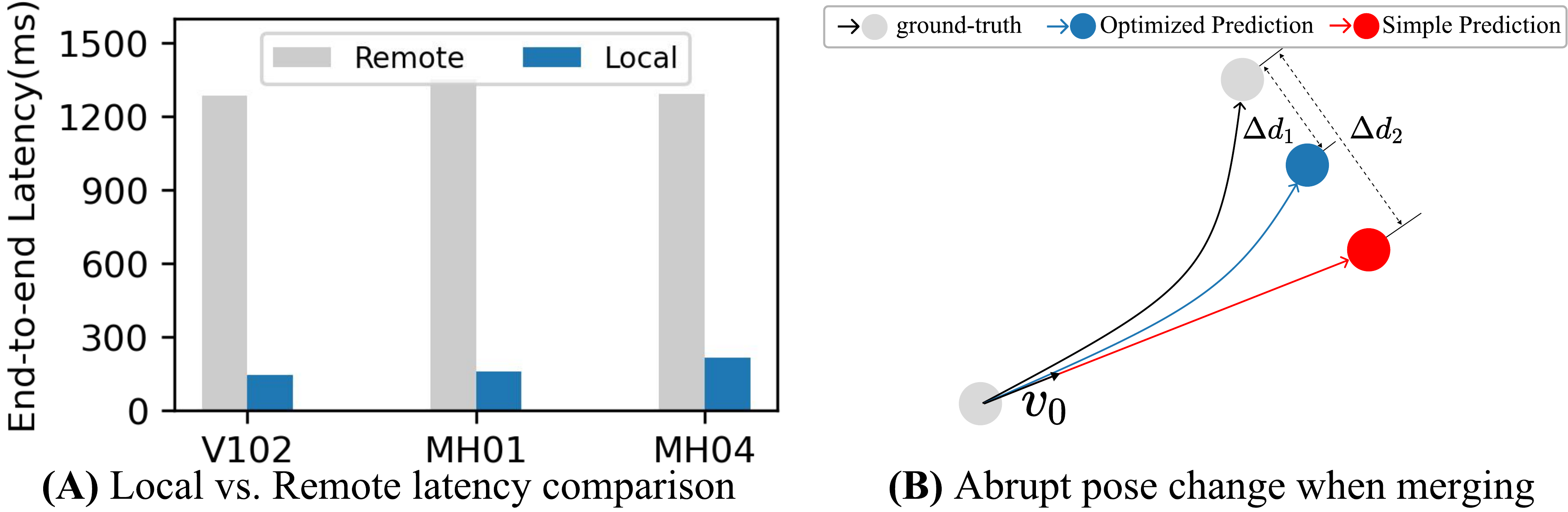}
\caption{Latency profiling and challenges during poses merge.}
\label{fig_exp}
\end{figure}

As shown in the Fig.~\ref{fig_exp}(A) gray bars, it usually takes more than 1.2 seconds for user to observe the feedback visualization of a transform invoked by the user motion in XR. This large latency causes large user-agent motion mismatch and makes the user adjust based on the old poses from agent, which results in false adjustment. We find out that the dominant part of the latency is due to the network transmission, which can go up to 1 second since the data has to suffer from a two-way network delay. 

Following this observation, we believe the timely update of agent poses is more critical compared to pose accuracy. Therefore, we opt for a duo-virtual spaces design which aims to avoid the network latency with a localization method running directly on XR. Since XR and agent are equipped with similar localization sensors like IMU and CAM, XR can set up models for agent and objects within the user-side virtual space and predict the poses using the models and directly localize the poses with the XR sensors. The agent keeps a second virtual space with the original user poses from XR, agent, and objects poses directly captured from agent sensors and uses this as a ground-truth state. Agent sends back the poses periodically to calibrate the state in user-side virtual space, which can reduce the positional error caused by prediction. We then make changes to the existing framework used in the previous experiment and record the end-to-end latency in Fig.~\ref{fig_exp} blue bars. We find out the end-to-end latency is reduced by 89\%, 88\%, and 83\% respectively for all three trajectories. 

However, this method faces significant pose drift when merging the ground-truth poses with the predicted poses for each of the calibrations, as shown in Fig.~\ref{fig_exp}(B). If the prediction does not optimize for a smooth transition for the next state, the object will create a large euclidean distance $\Delta d_2$ compared to the optimized predicted pose euclidean distance $\Delta d_1$. Solving this issue is our ongoing work.



\section{Conclusion}

This extended abstract describes a new system design for existing XR teleoperation aiming to reduce the large network latency between XR and agent. We profiled the end-to-end latency and provided a case study to prove the effectiveness of the new design. We plan to continue exploring further on challenges to complete this framework and conduct extensive experiments in various network and scene conditions.

\bibliography{ref}
\bibliographystyle{IEEEtran}

\end{document}